\pdfoutput=1

\documentclass[11pt]{article}

\usepackage{acl}

\usepackage{times}
\usepackage{latexsym}

\usepackage[T1]{fontenc}

\usepackage[utf8]{inputenc}

\usepackage{microtype}
\usepackage{hyperref}

\usepackage{booktabs}
\usepackage{graphicx}

\usepackage{amsmath}
\usepackage{xspace}
\usepackage{multirow}
\usepackage{array}

\newcommand{\bftab}[1]{\fontseries{b}\selectfont #1}

\newcolumntype{P}[1]{>{\centering\arraybackslash}p{#1}}

\begin{document}
\title{Zero and Few-shot Learning for Author Profiling}

\author{
Mara Chinea-Rios,
Thomas M{\"u}ller, \\
\textbf{Gretel Liz De la Peña Sarracén,
Francisco Rangel, \and
Marc Franco-Salvador}  \\
\vspace{-6pt}\\
Symanto Research, Valencia, Spain\phantom{$^1$ } \\
\url{https://www.symanto.com}\\
{\small \texttt{\{mara.chinea,thomas.mueller,gretel.delapena,francisco.rangel,marc.franco\}@symanto.com}}
}

\maketitle              

\begin{abstract}
Author profiling classifies author characteristics by analyzing how language is shared among people.
In this work, we study that task from a low-resource viewpoint: using little or no training data.
We explore different zero and few-shot models based on entailment and evaluate our systems on several profiling tasks in Spanish and English.
In addition, we study the effect of both the entailment hypothesis and the size of the few-shot training sample. 
We find that entailment-based models out-perform supervised text classifiers based on roberta-XLM
and that we can reach 80\% of the accuracy of previous approaches using less than 50\% of the training data on average.
\end{abstract}


\section{Introduction} \label{sec:introduction}

Author profiling \cite{rangel2013overview} aims to identify the characteristics or traits of an author by analyzing its sociolect, i.e., how language is shared among people.
It is used to determine traits such as age, gender, personality, or language variety.
The popularity of this task has increased notably in the last years.\footnote{According to \url{https://app.dimensions.ai/}, the term ``Author profiling'' tripled its frequency in research publications in the last ten years.}
Since 2013, the \textit{Uncovering Plagiarism Authorship and Social Software Misuse}\footnote{\url{https://pan.webis.de/}} (PAN) Lab at CLEF conference 
 organized annual shared tasks focused on different traits, e.g. age, gender, language variety, bots, or hate speech and fake news spreaders.

Data annotation for author profiling is a challenging task.
Aspects such as the economic and temporal cost, the psychological and linguistic expertise needed by the annotator, and the congenital subjectivity involved in the annotation task, make it difficult to obtain large amounts of high quality data \cite{bobicev-sokolova-2017-inter,troiano-etal-2021-emotion}. 
Furthermore, with few exceptions for some tasks (e.g. PAN), most of the research has been conducted on English corpora,
while other languages are under-resourced.

Few-shot (FS) learning aims to train classifiers with little training data.
The extreme case, zero-shot (ZS) learning \cite{larochelle2008,Chang2008ImportanceOS}, does not use any labeled data.
This is usually achieved by representing the labels of the task in a textual form, which can either be the name of the label or a concise textual description.
One popular approach \cite{yin-etal-2019-benchmarking,yin-etal-2020-universal,halder-etal-2020-task,Wang2021EntailmentAF} is based on textual entailment.
That task, a.k.a. Natural Language Inference (NLI) \cite{10.1007/11736790_9,bowman-etal-2015-large}, aims to predict whether a textual premise implies a textual hypothesis in a logical sense, e.g. \emph{Emma loves cats} implies that \emph{Emma likes cats}. 
The entailment approach for text classification uses the input text as the premise, and the text representing the label as the hypothesis. 
An NLI model is then applied to each premise-hypothesis pair, and the entailment probability is used to get the best matching label.


Our contributions are as follows:
(i) We study author profiling from a low-resourced viewpoint and explore different zero and few-shot models based on textual entailment. 
This is, to the best of our knowledge, the first attempt to use zero and few-shot learning for author profiling.
(ii) We study the identification of age, gender, hate speech spreaders, fake news spreaders, bots, and depression, in English and Spanish.
(iii) We analyze the effect of the entailment hypothesis and the size of the few-shot training sample on the system's performance. 
(iv) Our novel \textit{author instance selection} method allows to identify the most relevant texts of each author. 
Our experiments show that on average we can reach $80\%$ of the accuracy of the top PAN systems using less than $50\%$ of the training data.

\section{Zero and Few-shot Author Profiling Approach}
\label{sec:method}

In this section, we give details zero and few-shot text classification based on textual entailment and describe how we apply it to author profiling.

\subsection{Zero and Few-shot Text Classification}

We follow the so called entailment approach \cite{yin-etal-2019-benchmarking,yin-etal-2020-universal,halder-etal-2020-task,Wang2021EntailmentAF} to zero-shot text classification.
The approach relies on the capabilities of neural language models such as BERT \cite{devlin-etal-2019-bert} trained on large NLI datasets such as Stanford NLI \cite{bowman-etal-2015-large}.
These models are trained to predict whether a hypothesis text follows from a premise text.
The entailment approach consists of using the input text of a classification problem as the premise.
A hypothesis in textual form is then defined for each label.
The model is applied to each text-hypothesis pair and the label with the highest entailment probability is chosen as the predicted class.

To use this approach in a few-shot setup, we need to create a training set.
We follow the related work~\cite{Wang2021EntailmentAF} and generate the examples in the following manner:
Given an text instance and its reference label, we create an instance of class \emph{entailed} using the premise of the hypothesis of the reference label.
For every other label hypothesis, we create an instance of class \emph{contradicted}.

The neural models used for entailment classification are usually based on cross attention which means that both inputs (premise and hypothesis) are encoded jointly and their tokens can attend to each other.
However, recent work \cite{chu2020,fsl-siamese} has shown that Siamese networks can also work well if pre-trained on NLI data.
A Siamese network encodes premise and hypotheses into a high dimensional vector space and uses a similarity function such as the dot-product to find the hypothesis that best matches the input text.
These model can also be used in a few-shot set up. Here we follow the common approach \cite{fsl-siamese} of using the so called \emph{batch softmax} \cite{HendersonASSLGK17}:
\[
\mathcal{J} = - \frac{1}{B} \sum_{i=1}^{B} \left[ S(x_i, y_i) - \log \sum_{j=1}^{B} \exp S(x_i, y_j) \right]
\]
Where $B$ is the batch size and $S(x,y) = f(x) \cdot f(y)$ the similarity between input $x$ and label text $y$ under the current model $f$.
All other elements of the batch are used as \emph{in-batch negatives}.
We construct the batches so that every batch contains exactly one example of each label.

\subsection{Author Profiling}
\label{sec:method:ap}

Author profiling represents each author as a collection of texts. 
The objective is to assign each author to a label of a given set.
As the transformer models we employ in this study are large and scale quadratically with the input length we process each text as a separate instance.
Following the literature \cite{franco2015language}, we determine an author's class $y$ in function of the probabilities of
 classification of its texts: 
 \[
 y=\text{argmax}_{c \in C}\sum_{i=1}^n \mathrm{P}(c \mid t_i)
 \]
 Where $C$ is the total number of classes and $[t_1,...,t_n]$ is the list of texts of that specific author.

Typically, the labeled texts of an author are random and might thus not be representative of the author's label.
Especially, when separating the texts into individual training instances, this might produce noisy or misleading instances.
In this work, we propose an \textit{instance selection} method to mitigate this effect.
Given the texts $T$, it returns the set of texts $T' = \{t_1, ..., t_m\}$ where each text $t_i$ is the closest to the $i^\text{th}$ centroid $d_i \in D$.
The cluster set $D$ results of applying the Agglomerative Clustering method with cosine distance to the texts in $T$.
Following other work\footnote{\url{https://tinyurl.com/st-agg-cluster}}, we use scikit-learn \cite{scikit-learn} and a distance threshold\ of 1.5 to get a dynamic number of clusters depending on the author and its information.  
\section{Related Work}
\label{sec:rel_work}


Early \textit{Author Profiling} attempts focused on blogs and formal text~\cite{argamon:2003,koppel:2003} based on Pennebaker's~\cite{pennebaker:2003} 
theory, which connects the use of the language with the personality traits of the authors. 
With the rise of social media, researchers proposed methodologies to profile the authors of posts where the language is more informal~\cite{burger:2011}.
Since then, several approaches have been explored. 
For instance, based on second order representation which relates documents and user profiles~\cite{pastor:2013}, the Emograph graph-based approach enriched with topics and emotions~\cite{rangel:ipm:2016}, or the LDSE~\cite{rangel:2016:cicling}, commonly used as a baseline at PAN.
Recently, the research has focused on the identification of bots and users who spread harmful information (e.g. fake news and hate speech).
In addition, there has been work to leverage the impact of personality traits and emotions to discriminate between classes~\cite{ghanem2019emotional}. 

\textit{Zero and Few-shot Text Classification} 
has been explored in different manners. 
Semantic similarity methods use the explicit meaning of the label names to compute the similarity with the input text \cite{reimers-2019-sentence-bert}.
Prompt-based methods \cite{schick-schutze-2021-exploiting} use natural language generation models, such as GPT-3 \cite{NEURIPS2020_1457c0d6}, to get the most likely label to be associated with the input text.
In this work, we use entailment methods \cite{yin-etal-2019-benchmarking,halder-etal-2020-task}.
Recently, Siamese Networks have been found to give similar accuracy while being much faster \cite{fsl-siamese}.

\section{Experimental setup} \label{sec:experiments-setup}
\subsection{Dataset}
\begin{table*}[!tp]
\centering
\resizebox{1.0\textwidth}{!}{%
\begin{tabular}{ccc@{\hskip 0.5cm}c}
\toprule
 task & language & train & test \\
\midrule
\multirow{2}{*}{Gender \cite{rangel2017gender}} & EN & female:1427, male:1453 &  female:1200, male:1200 \\
& ES & female:1681, male:1679 & female:1400, male:1400\\
\midrule
\multirow{2}{*}{Age \cite{rangel2015age} } & EN & 18-24:58, 25-34:60, 35-49:22, 50+:12 & 18-24:56, 25-34:58, 35-49:20, 50+:8\\
& ES & 18-24:22, 25-34:46, 35-49:22, 50+:10 & 18-24:18, 25-34:44, 35-49:18, 50+:8 \\
\midrule
\multirow{2}{*}{Hate Speech \cite{rangel2021hatespeech}} & EN & hate:100,  not-hate:100 & hate:50, not-hate:50  \\
& ES & hate:100, not-hate:100 & hate:50, not-hate:50  \\
\midrule
\multirow{2}{*}{Bots \cite{rangel2019bots}} & EN & bot:2060, human:2060 & bot:1320, human:1320  \\
& ES & bot:1500, human:1500 & bot:900, human:900  \\
\midrule
\multirow{2}{*}{Fake News \cite{rangel2020fake}} & EN & spreader:150,not-spreader:150 & spreader:100, not-spreader:100\\
& ES & spreader:150, not-spreader:150 & spreader:100, not-spreader:100\\
\midrule
Depression \cite{parapar2021erisk} & EN & \texttt{--}:14, \texttt{-}:27, \texttt{+}:27, \texttt{++}:22 & \texttt{--}:6, \texttt{-}:34, \texttt{+}:27, \texttt{++}:13\\
\bottomrule
\end{tabular}}
\caption{Dataset overview. train and test show the number of users for each class. The depression categories are minimal (\texttt{--}), mild (\texttt{-}), moderate (\texttt{+}) and severe (\texttt{++}).}
\label{tab:dataset_statistics}
\end{table*}

We conduct a comprehensive study in 2 languages (English and Spanish) and 7 author profiling tasks:    
demographics (gender, age), hate-speech spreaders, bot detection, fake news spreaders, and depression level. 
We use datasets from the PAN and \textit{early risk prediction on the internet} (eRisk)~\cite{parapar2021erisk} shared tasks.
Table \ref{tab:dataset_statistics} gives details on the datasets.

\subsection{Entailment Models for Zero and Few Shot}

In our experiments, we use pretrained models hosted on Hugging Face \cite{wolf-etal-2020-transformers}.
Based on our prototyping experimentation, for the Cross Attention (CA) models, we use a \textit{BART large} \cite{nie2019adversarial}  model\footnote{\url{https://tinyurl.com/bart-large-snli-mnli}} for English and a \textit{XLM roberta-large}\cite{liu2019roberta} model \footnote{\url{https://tinyurl.com/xlm-roberta-large-xnli-anli}} for Spanish.
Following \cite{fsl-siamese}, for the Siamese Networks (SN) model, we use \textit{paraphrase-multilingual-mpnet-base-v2}\footnote{\url{https://tinyurl.com/paraphrase-multilingual-mpnet}}, a sentence transformer model \cite{reimers-2019-sentence-bert}, for English and Spanish.
All models have been trained on NLI data. 
The SN models has additionally been trained on paraphrase data.

\subsection{Baseline and Compared Approaches} \label{sec:baselines} 

 \begin{itemize}
 \item \textit{Best performing system} (winner): 
 We show the test results of the system that ranked first at each shared task overview paper.
 The reference to those specific systems can be found in the Appendix (S. \ref{sec:appendix}).

 \item \textit{Sentence-Transformers with logistic regression} (ST-lr): 
 We use the scikit-learn logistic regression classifier \cite{scikit-learn} on top of the embeddings of the Sentence Transformer model used for the Siamese Network experiments.
 
 \item \textit{Character $n$-grams with logistic regression} (user-char-lr):
 We use $[1..5]$ character $n$-grams with a TF-IDF weighting calculated using all texts.
 Note that $n-grams$ and logistic regression are predominant among the best systems in the different PAN shared tasks \cite{rangel2019bots,rangel2020fake}. 
 
  \item \textit{XLM-RoBERTa} (xlm-roberta): 
  Model based on the pre-trained \textit{XLM roberta-base} \cite{liu2019roberta}.
  Trained for 2 epochs, batch size of 32, and learning rate of 2e-4. 
 
 \item \textit{Low-Dimensionality Statistical Embedding} (LDSE): 
 This method \cite{rangel:2016:cicling} represents texts based on the probability distribution of the occurrence of their terms in the different classes. 
 Key of LDSE is to weight the probability of a term to belong to each class.
 Note that this is one of the best ranked baselines at different PAN editions.
 \end{itemize}

\subsection{Methodology and Parameters}
\label{eval:params}

We conduct our validation experiments using 5-fold cross-validation. 
We report the mean and standard deviation among folds.
Using this scheme we fine-tune the few-shot models in terms of users per label ($n \in [8,16,32,48,64,128,256,512]$), best entailment hypothesis, learning rate, and user instance selection method.
Following the literature~\cite{fsl-siamese}, SN uses a batch size equal to the number of labels and trains for 10 epochs.
CA uses a batch size of 8 and trains for 10 epochs.

We compare our author Instance Selection ($IS$) method (Section~\ref{sec:method:ap}) with two baseline methods that respectively select 1 and 50 random instances from the author text list ($Ra_1$ and $Ra_{50}$). 
We use the \textit{paraphrase-multilingual-mpnet-base-v2} model to obtain the instance embeddings required by $IS$.
The resulting tuned CA learning rates are 1e-8 and 1-e6 for $Ra_x$ and $IS$, respectively. For SN they are 2e-5 and 2e-6.

We use macro F1-score ($F_{1}$) as our main evaluation metric and accuracy ($Acc.$) to compare with the official shared task results.

\subsection{Hypotheses of the Entailment Models} 
\begin{table*} [!t]
	\centering
	
    \resizebox{1.0\textwidth}{!}{
    \begin{tabular}{p{1.8cm}p{0.8cm}p{0.5cm}p{4.8cm}p{0.5cm}p{4.8cm}}
	\toprule
		 & & \multicolumn{2}{c}{English} & \multicolumn{2}{c}{Spanish} \\
		Task & model & \#EN & Per-class hypothesis list & \#ES & 
		Per-class hypothesis list \\
		\midrule
		\multirow{2}{*}{Gender} & CA & $8$ & I'm a \{female, male\} &  $4$ &\textit{identity hypothesis} \\
		& SN & $8$ & My name is \{Ashley, Robert\} & $4$ & Soy \{una mujer, un hombre\} \\
		\midrule
		\multirow{5}{*}{Age} & CA & $6$ & \textit{identity hypothesis} & $5$ & La edad de esta persona es \{entre 18 y 24, entre 25 y 34, entre 35 y 49, más de 50\} años \\
		& SN  & $6$ & I am \{a teenager, a young adult, an adult, middle-aged\} & $5$ & \textit{identity hypothesis} \\
		\midrule
		\multirow{7}{*}{Hate Speech} & CA & $8$ & This text expresses prejudice and hate speech; This text does not contain hate speech & $6$ & Este texto expresa odio o prejuicios; Este texto es moderado, respetuoso, cortés y civilizado\\
		& SN & $8$ & This text contains prejudice and hate speech directed at particular groups; This text does not contain hate speech & $6$ & Odio esto; Soy respetuoso \\ 
		\midrule
		\multirow{3}{*}{Bots} & CA & $7$ & This is a text from a machine; This is a text from a person & $6$ & \textit{identity hypothesis} \\
		& SN & $7$ & \textit{identity hypothesis} & $6$ & Soy un bot; Soy un usuario\\
	    \midrule
	    \multirow{5}{*}{Fake News} & CA & $8$ & This author spreads fake news; This author is a normal user & $6$ & Este autor publica noticias falsas; Este autor es un usuario normal \\
		& SN & $8$ & \textit{identity hypothesis} & $6$ & Este usuario propaga noticias falsas; Este usuario no propaga noticias falsas\\
		\midrule
	    \multirow{3}{*}{Depression}	& CA & $4$ & The risk of depression of this user is \{minimal,mild, moderate, severe\} & \\
		& SN & $4$ & \textit{identity hypothesis} & \\
		\bottomrule
	\end{tabular} }
	\caption{\label{tab:Hypothesis-table} Best performing entailment hypotheses. \#\{EN,ES\} indicates the number of explored hypotheses per language.}
\end{table*}

We compared 68 different hypotheses for the CA and SN entailment models. We included the \textit{identity hypothesis}: represent the label using its raw string label. 
Table \ref{tab:Hypothesis-table}
shows the best performing hypothesis for each task and model.
We use those for the rest of the evaluation.
See the Appendix (S. \ref{sec:appendix}) for a complete list with results of all the hypotheses explored in our experiments.

\section{Results and Analysis} \label{sec:eval}

\begin{table*} [!tp]
	\centering
	\resizebox{1.0\textwidth}{!}{
	\begin{tabular}{lllllllllllllll}
	\toprule
		&&& \multicolumn{2}{c}{Gender} & \multicolumn{2}{c}{Age} & 
		\multicolumn{2}{c}{Hate Speech} & 
		\multicolumn{2}{c}{Bots} & 
		\multicolumn{2}{c}{Fake News} & 
		\multicolumn{2}{c}{Depression} \\
		\cmidrule(lr){4-5} \cmidrule(lr){6-7} \cmidrule(lr){8-9}  \cmidrule(lr){10-11}
		 \cmidrule(lr){12-13} \cmidrule(lr){14-15}
		& \textit{n} & Inst. Sel. & \textit{s} & $F_{1}$ & s & $F_{1}$ & s & $F_{1}$ & s & $F_{1}$ & s & $F_{1}$ & s & $F_{1}$\\
	   \midrule
	   \parbox[t]{5mm}{\multirow{25}{*}{\rotatebox[origin=c]{90}{Cross-Attention (CA)}}}
 		& $0$ & - & $0$ & $74.7_{1.5}$ & $0$ & $27.6_{2.6}$ & $0$ & \bftab{71.0$_{2.8}$} & $0$ & $42.4_{1.2}$ & $0$ & $63.8_{3.7}$ & $0$ & $24.1_{2.3}$\\
        \cmidrule{2-15}
	    & \multirow{3}{*}{$8$} & $Ra_{1}$ & $16$ & $ 74.9_{1.5}$ & $32$ & $27.4_{3.0}$ & $16$ & $68.8_{4.4}$ & $16$ & $74.4_{11.5}$ & $16$ & $62.3_{5.2}$ & $29$ & $27.6_{1.9}$\\
		& & $Ra_{50}$ & $800$ & $73.2_{2.7}$ & $1057$ & $14.9_{4.0}$ & $800$ & $45.9_{6.9}$ & $800$ & $86.3_{1.4}$ & $800$ & $57.7_{8.3}$ & $1,3k$ & $20.5_{6.5}$\\
		& & $IS$ & $257$ & $67.5_{4.3}$ &$361$ & $34.6_{3.3}$ & $348$ & $60.1_{7.2}$ & $191$ & $87.6_{1.4}$ & $203$ & $54.8_{3.1}$ & $1,9k$ & $23.6_{10.9}$\\
		\cmidrule{2-15}
 		& \multirow{3}{*}{$16$} & $Ra_{1}$ & $32$ & $74.9_{1.5}$ & $46$ & $27.4_{2.4}$ & $32$ & $66.1_{3.8}$ & $32$ & $79.2_{6.2}$ & $32$ & $62.6_{4.1}$ & $45$ & \bftab{27.8$_{2.0}$} \\
		& & $Ra_{50}$ & $1,6k$ & $67.7_{5.3}$ & $2,3k$ & $12.5_{0.6}$ & $1,6k$ & $45.5_{5.0}$ & $1,6k$ & $87.1_{0.4}$ & $1,6k$ & $55.9_{8.5}$ & $2,2k$ & $ 13.6_{6.4}$ \\
		& & $IS$ & $501$ & $71.3_{3.6}$ &$611$ & $37.1_{6.1}$ & $697$ & $59.6_{6.0}$ & $385$ & $88.5_{1.2}$ & $432$ & $62.3_{4.9}$ & $2,9k$ & $21.4_{6.7}$\\
        \cmidrule{2-15}
        & \multirow{3}{*}{$32$} & $Ra_{1}$ & $64$ & $74.8_{1.5}$ & $78$ & $27.2_{2.3}$ & $64$ & $65.9_{5.5}$ & $64$ & $86.3_{3.3}$ & $64$ & $65.9_{5.5}$ & - & - \\
		& & $Ra_{50}$ & $3,2k$ & $69.2_{4.1}$ & $3,9k$ & $12.5_{0.6}$ & $3,2k$ & $56.2_{10.3}$ & $3,2k$ & $88.8_{1.3}$ & $3,2k$ & $48.4_{2.8}$ & - & - \\
 		& & $IS$ & $1,0k$ & $75.5_{3.4}$ &$1,0k$ & \bftab{41.7$_{3.4}$} & $1,4k$ & $56.0_{6.5}$ & $756$ & $89.7_{1.2}$ & $880$ & $61.3_{4.5}$ & - & -\\
        \cmidrule{2-15}
        & \multirow{3}{*}{$48$} & $Ra_{1}$ & $96$ & $74.8_{1.2}$ & - & - & $96$ & $62.2_{8.1}$ & $96$ & $89.9_{1.2}$ & $96$ & $65.5_{3.5}$ & - & - \\
        & & $Ra_{50}$ & $4,8k$ & $72.5_{1.9}$ & - & - & $4,8k$ & $56.1_{10.3}$ & $4,8k$ & $90.5_{0.9}$ & $4,8k$ & $50.0_{4.5}$ & - & - \\
        & & $IS$ & $1517$ & $75.0_{2.6}$ & - & - & $2059$ & $66.4_{11.0}$ & $1132$ & $90.4_{1.4}$ & $1320$ & \bftab{65.8$_{5.3}$} & - & -\\
        \cmidrule{2-15}
        & \multirow{3}{*}{$64$} & $Ra_{1}$ & $128$ & $74.5_{1.2}$ & - & - & $128$ & $59.0_{10.5}$ & $128$ & $90.7_{1.2}$ & $128$ & $65.3_{3.3}$ & - & - \\
        & & $Ra_{50}$ & $6,4k$ & $74.8_{2.2}$ & - & - & $6,4k$ & $62.9_{14.3}$ & $6,4k$ & $90.1_{2.1}$ & $6,4k$ & $53.0_{2.6}$ & - & - \\
        & & $IS$ & $2,0k$ & $75.6_{2.7}$ & - & - & $2,2$ & $59.5_{8.7}$ & $1,5k$ & $90.4_{1.4}$ & $1,8k$ & $61.2_{4.1}$ & - & -\\
		\cmidrule{2-15}
        & \multirow{3}{*}{$128$} & $Ra_{1}$ & $256$ & $74.9_{0.8}$ & - & - & - & - & $256$ & $90.4_{2.0}$ & - & - & - & - \\
		& & $Ra_{50}$ & $12,8k$ & $74.3_{1.6}$ & - & - & - & - & $12,8k$ & $92.3_{1.9}$ & - & - & - & - \\
		& & $IS$ & $4,0k$ & $76.4_{1.7}$ & - & - & - & - & $3,0k$ & $92.1_{1.6}$ & - & - & - & -\\
		\cmidrule{2-15}
        & \multirow{3}{*}{$256$} & $Ra_{1}$ & $512$ & $75.6_{1.3}$ & - & - & - & - & $512$ & $91.3_{0.8}$ & - & - & - & - \\
		& & $Ra_{50}$ & $25,6k$ & $73.9_{0.9}$ & - & - & - & - & $25,6k$ & $95.8_{1.3}$ & - & - & - & - \\
 		& & $IS$ & $8,0k$ & \bftab{79.0$_{1.2}$} & - & - & - & - & $6,0k$ & $94.5_{1.4}$ & - & - & - & -\\
		\cmidrule{2-15}
        & \multirow{3}{*}{$512$} & $Ra_{1}$ & $1,0k$ & $70.9 _{1.3}$ & - & - & - & - & $1,0k$ & $93.0_{1.1}$ & - & - & - & - \\
 		& & $Ra_{50}$ & $51,2k$ & $74.3_{0.6}$ & - & - & - & - & $51,2k$ & \bftab{97.9$_{0.6}$} & - & - & - & - \\
		& & $IS$ & $16,0k$ & $77.5_{2.3}$ & - & - & - & - & $11,9k$ & $97.0_{0.6}$ & - & - & - & -\\
		\toprule
	    \parbox[t]{5mm}{\multirow{25}{*}{\rotatebox[origin=c]{90}{Siamese Network (SN)}}}
	    & $0$ & - & $0$ & $38.1_{3.5}$ & $0$ & $37.4_{6.6}$ & $0$ & $46.3_{12.4}$ & $0$ & $62.2_{1.4}$ & $0$ & $51.8_{2.4}$ & $0$ & $21.9_{5.2}$ \\
	    \cmidrule{2-15}
		&\multirow{3}{*}{$8$} & $Ra_{1}$ & $16$ & $55.1_{8.7}$ & $32$ & $39.6_{10.9}$ & $16$ & $41.7_{6.1}$ & $16$ & $50.4_{10.1}$ & $16$ & $58.8_{7.9}$ &  $29$ & $26.3_{5.5}$\\
		&& $Ra_{50}$ & $800$ & $63.1_{6.4}$ & $1057$ & $50.5_{7.1}$ & $800$ & $62.6_{7.8}$ & $800$ & $84.5 _{1.3}$ & $800$ & $55.8_{5.6}$ & $1,4k$ & $29.2_{3.5}$\\
 		&& $IS$ & $257$ & $65.2_{4.8}$ & $361$ & $51.7_{7.5}$ & $348$ & $60.9_{3.7}$ & $191$ & $86.1_{1.8}$ & $203$ & $54.0_{5.5}$ & $1,9k$ & \bftab{32.7$_{7.2}$}\\
		\cmidrule{2-15}
		&\multirow{3}{*}{$16$} & $Ra_{1}$ & $32$ & $59.4_{3.2}$ & $46$ & $39.6_{7.1}$ & $32$ & $ 47.1_{7.1}$ & $32$ & $73.3_{8.9}$ & $32$ & $59.5_{3.0}$ & $45$ & $23.7_{5.5}$ \\
        && $Ra_{50}$ & $1,6k$ & $68.2_{4.1}$ & $2,3k$ & $55.4_{4.2}$ & $1,6k$ & $59.5_{4.7}$ & $1,6k$ & $84.7_{2.8}$ & $1,6k$ & $58.2_{7.4}$ & $2,2k$ & $25.1 _{7.1}$ \\
        && $IS$ & $501$ & $69.3_{3.6}$ &$611$ & $50.9_{4.2}$ & $697$ & $60.7_{9.4}$ & $385$ & $86.9_{1.8}$ & $432$ & $61.7_{7.4}$ & $2,9k$ & $28.8_{4.8}$\\
	    \cmidrule{2-15}
		&\multirow{3}{*}{$32$} & $Ra_{1}$ & $64$ & $59.6_{5.9}$ & $78$ & $38.5_{7.1}$ & $64$ & $49.0_{4.9}$ & $64$ & $84.5_{1.8}$ & $64$ & $60.4_{3.8}$ & - & - \\
        && $Ra_{50}$ & $3,2k$ & $71.6_{2.3}$ & $3,9k$ & \bftab{52.9$_{7.4}$} & $3,2k$ & $58.8_{5.2}$ & $3,2k$ & $87.1_{4.1}$ & $3,2k$ & $65.2_{5.7}$ & - & - \\
        && $IS$ & $1,0k$ & $71.4_{2.7}$ &$1,0k$ & $51.4_{8.0}$ & $1,4k$ & $60.4_{5.3}$ & $756$ & $88.6_{1.6}$ & $880$ & $62.6_{5.5}$ & - & -\\
		\cmidrule{2-15}
		&\multirow{3}{*}{$48$} & $Ra_{1}$ & $96$ & $65.6_{3.8}$ & - & - & $96$ & $57.9_{7.2}$ & $96$ & $86.5_{2.9}$ & $96$ & $ 63.3_{3.3}$ & - & - \\
        && $Ra_{50}$ & $4,8k$ & $73.1_{1.6}$ & - & - & $4,8k$ & $65.5_{4.9}$ & $4,8k$ & $89.6_{1.4}$ & $4,8k$ & $65.5_{4.9}$ & - & - \\
        && $IS$ & $1,5k$ & $72.6_{2.4}$ & - & - & $2,0k$ & $67.8_{4.1}$ & $1,1k$ & $90.1_{1.7}$ & $1,3k$ & \bftab{68.4$_{5.8}$} & - & -\\
	    \cmidrule{2-15}
		&\multirow{3}{*}{$64$} & $Ra_{1}$ & $128$ & $66.5_{4.2}$ & - & - & $128$ & $56.3_{9.5}$ & $128$ & $88.2_{1.5}$ & $128$ & $56.3_{9.5}$ & - & - \\
        && $Ra_{50}$ & $6,4k$ & $74.1_{2.1}$ & - & - & $6,4k$ & $68.7_{6.0}$ & $6,4k$ & $89.6_{1.5}$ & $6400$ & $63.9_{4.5}$ & - & - \\
        && $IS$ & $2,0k$ & $74.4_{2.6}$ & - & - & $2,2k$ & \bftab{68.4$_{5.9}$} & $1,5k$ & $89.8_{1.5}$ & $1,8k$ & $63.0_{6.2}$ & - & -\\
        \cmidrule{2-15}
		&\multirow{3}{*}{$128$} & $Ra_{1}$ & $256$ & $67.8 _{2.3}$ & - & - & - & - & $256$ & $89.8_{1.9}$ & - & - & - & - \\
        && $Ra_{50}$ & $12,8k$ & $75.8_{0.7}$ & - & - & - & - & $12,8k$ & $92.4_{1.7}$ & - & - & - & - \\
        && $IS$ & $4,0k$ & $75.3_{0.6}$ & - & - & - & - & $3,0k$ & $92.5_{1.7}$ & - & - & - & -\\
		\cmidrule{2-15}
		&\multirow{3}{*}{$256$} & $Ra_{1}$ & $512$ & $70.3_{1.9}$ & - & - & - & - & $512$ & $91.2_{1.6}$ & - & - & - & - \\
		&& $Ra_{50}$ & $25,6k$ & $77.1_{1.5}$ & - & - & - & - & $25,6k$ & $95.4_{1.7}$ & - & - & - & - \\
		&& $IS$ & $8,0k$ & $76.4_{1.4}$ & - & - & - & - & $6,0k$ & $94.5_{2.2}$ & - & - & - & -\\
		\cmidrule{2-15}
	    &\multirow{3}{*}{$512$} & $Ra_{1}$ & $1,0k$ & $73.9 _{1.9}$ & - & - & - & - & $1,0k$ & $93.4_{1.7}$ & - & - & - & - \\
 		&& $Ra_{50}$ & $51,2k$ & $77.1_{1.7}$ & - & - & - & - & $51,2k$ & \bftab{97.8$_{0.9}$} & - & - & - & - \\
 		&& $IS$ & $15,9k$ & \bftab{77.1$_{2.0}$} & - & - & - & - & $11,8k$ & $96.9_{1.1}$ & - & - & - & -\\
	     \bottomrule
	\end{tabular}}
	\caption{\label{tab:comparative_train_creation_instance_user_english} \textbf{English Validation results} of the CA and SN models. \textit{n} shows the number of training users and \textit{s} the total training size. Top results are highlighted with \textbf{bold}.}
\end{table*}

\begin{table*} [!tp]
	\centering
	\resizebox{1.0\textwidth}{!}{
	\begin{tabular}{llll@{\hskip 0.5cm}ll@{\hskip 0.5cm}ll@{\hskip 0.5cm}ll@{\hskip 0.5cm}ll@{\hskip 0.5cm}l}
	\toprule
		&& & \multicolumn{2}{c}{Gender} & \multicolumn{2}{c}{Age} & 
		\multicolumn{2}{c}{Hate Speech} & 
		\multicolumn{2}{c}{Bots} & 
		\multicolumn{2}{c}{Fake News} \\
		\cmidrule(lr){4-5} \cmidrule(lr){6-7} \cmidrule(lr){8-9}  \cmidrule(lr){10-11}
		\cmidrule(lr){12-13}
		&\textit{n} & Inst. Sel. & \textit{s} & $F_{1}$ & s & $F_{1}$ & s & $F_{1}$ & s & $F_{1}$ & s & $F_{1}$ \\
		\midrule
		\parbox[t]{5mm}{\multirow{25}{*}{\rotatebox[origin=c]{90}{Cross Attention (CA)}}}
		&$0$ & - & $0$ & $68.1_{2.6}$ & $0$ & \bftab{27.1$_{4.3}$} & $0$ & $60.5_{9.0}$ & $0$ & $36.1_{1.2}$ & $0$ & $36.3_{3.5}$ \\
        \cmidrule{2-13}
	    &\multirow{3}{*}{$8$} & $Ra_{1}$ & $16$ & $61.9_{11.3}$ & $26$ & $14.0_{2.3}$ & $16$ & $61.9_{11.3}$ & $16$ & $68.0_{4.1}$ & $16$ & $35.1_{2.5}$\\
		&& $Ra_{50}$ & $800$ & $ 61.3_{2.1}$ & $1,2k$ & $12.8_{0.4}$ & $800$ & $64.2_{7.5}$ & $800$ & $72.6_{5.3}$ & $800$ & $34.5_{2.6}$ \\
		&& $IS$ & $204$ & $64.0_{7.3}$ &$248$ & $24.1_{7.8}$ & $272$ & $63.2_{14.4}$ & $160$ & $77.0_{4.4}$ & $168$ & $66.0_{7.5}$ \\
		\cmidrule{2-13}
		&\multirow{3}{*}{$16$} & $Ra_{1}$ & $32$ & $67.7_{2.4}$ & $40$ & $14.0_{2.3}$ & $32$ & $61.1_{10.7}$ & $32$ & $72.1_{2.1}$ & $32$ & $35.1_{2.5}$ \\
		&& $Ra_{50}$ & $1,6k$ & $56.8_{13.1}$ & $1.9k$ & $12.8_{0.4}$ & $1,6k$ & $69.6 _{7.3}$ & $1,6k$ & $79.2_{4.8}$ & $1,6k$ & $45.3_{9.5}$ \\
		&& $IS$ & $401$ & $65.6_{3.8}$ & $474$ & $18.4_{8.1}$ & $542$ & $68.9_{7.7}$ & $326$ & $82.0_{2.9}$ & $334$ & $70.9_{4.5}$ \\
        \cmidrule{2-13}
        &\multirow{3}{*}{$32$} & $Ra_{1}$ & $64$ & $67.6_{2.1}$ & - & - & $64$ & $62.8_{12.1}$ & $64$ & $74.9_{4.0}$ & $64$ & $35.3_{2.5}$  \\
		&& $Ra_{50}$ & $3,2k$ & $59.9_{15.0}$ & - & - & $3,2k$ & $59.6_{7.9}$ & $3,2k$ & $82.9 _{2.8}$ & $3,2k$ & $63.5_{4.5}$ \\
		&& $IS$ & $807$ & $69.5_{1.6}$ & - & - & $1,9k$ & $75.8_{2.0}$ & $630$ & $84.7_{1.8}$ & $673$ & $73.6_{2.1}$ \\
        \cmidrule{2-13}
        &\multirow{3}{*}{$48$} & $Ra_{1}$ & $96$ & $67.3_{2.2}$ & - & - & $96$ & $63.7_{11.0}$ & $96$ & $77.4_{2.4}$ & $96$ & $35.3_{2.5}$ \\
        && $Ra_{50}$ & $4,8k$ & $61.2_{15.6}$ & - & - & $4,8k$ & $50.5_{6.8}$ & $4,8k$ & $84.8 _{4.2}$ & $4,8k$ & $65.2_{4.9}$\\
        && $IS$ & $1,2k$ & $69.1_{2.2}$ & - & - & $1,6k$ & $74.9_{5.0}$ & $954$ & $85.1_{3.9}$ & $1.0k$ & \bftab{77.5$_{3.3}$} \\
        \cmidrule{2-13}
        &\multirow{3}{*}{$64$} & $Ra_{1}$ & $128$ & $67.1_{2.0}$ & - & - & $128$ & $62.3_{9.5}$ & $128$ & $78.4_{2.5}$ & $128$ & $34.8_{2.6}$\\
        && $Ra_{50}$ & $6,4k$ & $61.4_{15.7}$ & - & - & $6,4k$ & $49.1_{4.7}$ & $6,4k$ & $86.1 _{2.5}$ & $6,4k$ & $67.4_{3.9}$\\
        && $IS$ & $1,6k$ & $69.5_{2.6}$ & - & - & $2,2k$ & \bftab{77.1$_{2.0}$} & $1,3k$ & $86.0_{2.2}$ & $1,3k$ & $75.0_{2.2}$\\
		\cmidrule{2-13}
        &\multirow{3}{*}{$128$} & $Ra_{1}$ & $256$ & $65.8_{2.2}$ & - & - & - & - & $256$ & $83.1_{2.0}$ & - & - \\
		&& $Ra_{50}$ & $12,8k$ & $55.0_{17.8}$ & - & - & - & - & $12,8k$ & $91.4_{1.7}$ & - & - \\
		&& $IS$ & $3,3k$ & $72.1_{2.0}$ & - & - & - & - & $2,6k$ & $89.9_{1.8}$ & - & -\\
		\cmidrule{2-13}
        &\multirow{3}{*}{$256$} & $Ra_{1}$ & $512$ & $63.4_{1.4}$ & - & - & - & - & $512$ & $87.1_{1.9}$ & - & - \\
		&& $Ra_{50}$ & $25,6k$ & $68.1_{2.6}$ & - & - & - & - & $25,6k$ & $95.3_{1.4}$ & - & - \\
		&& $IS$ & $5,2k$ & $70.4_{3.0}$ & - & - & - & - & $5,1k$ & $92.5_{1.5}$ & - & - \\
		\cmidrule{2-13}
        &\multirow{3}{*}{$512$} & $Ra_{1}$ & $1,0k$ & $61.5_{1.9}$ & - & - & - & - & $1,0k$ & $89.0_{1.2}$ & - & - \\
		&& $Ra_{50}$ & $51,2k$ & $68.1_{2.6}$ & - & - & - & - & $51,2k$ & \bftab{97.4$_{1.7}$} & - & - \\
		&& $IS$ & $13,0k$ & \bftab{72.5$_{2.0}$} & - & - & - & - & $10,3k$ & $94.9_{1.7}$ & - & - \\
		\toprule
  	    \parbox[t]{5mm}{\multirow{25}{*}{\rotatebox[origin=c]{90}{Siamese Network (SN)}}}
	    &$0$ & - & $0$ & $33.6_{1.3}$ & $0$ & $31.8_{3.6}$ & $0$ & $36.9_{3.9}$ & $0$ & $50.1_{1.6}$ & $0$ & $51.5_{6.7}$ \\
	    \cmidrule{2-13}
		&\multirow{3}{*}{$8$} & $Ra_{1}$ & $16$ & $50.1_{7.1}$ & $26$ & $31.8_{5.0}$ & $16$ & $54.4_{16.6}$ & $16$ & $68.8_{12.1}$ & $16$ & $64.0_{8.2}$\\
		&& $Ra_{50}$ & $800$ & $60.2_{6.7}$ & $1,2k$ & $51.0_{12.6}$ & $800$ & $70.3_{10.0}$ & $800$ & $76.4_{3.6}$ & $800$ & $71.7_{8.0}$ \\
		&& $IS$ & $204$ & $62.0_{6.9}$ &$248$ & $44.4_{12.0}$ & $272$ & $71.0_{12.5}$ & $160$ & $80.2_{4.9}$ & $168$ & $70.5_{5.0}$ \\
		\cmidrule{2-13}
		&\multirow{3}{*}{$16$} & $Ra_{1}$ & $32$ & $60.7_{4.3}$ & $40$ & $30.7_{9.6}$ & $32$ & $63.8_{8.2}$ & $32$ & $76.8_{7.7}$ & $32$ & $65.7_{6.6}$ \\
        && $Ra_{50}$ & $1,6k$ & $61.7_{2.2}$ & $2,0k$ & \bftab{53.0$_{10.2}$} & $1,6k$ & $71.0 _{7.0}$ & $1,6k$ & $78.3_{6.3}$ & $1,6k$ & $74.3_{3.7}$ \\
        && $IS$ & $401$ & $63.9_{1.7}$ & $474$ & $49.6_{12.7}$ & $542$ & $75.5_{4.6}$ & $326$ & $82.6_{3.8}$ & $334$ & $73.9_{2.4}$ \\
	    \cmidrule{2-13}
	    &\multirow{3}{*}{$32$} & $Ra_{1}$ & $64$ & $64.1_{2.7}$ & - & - & $64$ & $68.8_{7.5}$ & $64$ & $82.1_{1.4}$ & $64$ & $71.3_{7.1}$  \\
		&& $Ra_{50}$ & $3,2k$ & $66.7_{2.7}$ & - & - & $3,2k$ & $74.9_{3.6}$ & $3,2k$ & $74.9_{3.6}$ & $3,2k$ & $74.9_{3.6}$ \\
		&& $IS$ & $807$ & $67.1_{1.6}$ & - & - & $1,9k$ & $78.1_{4.6}$ & $630$ & $85.3_{3.5}$ & $673$ & $75.0_{2.6}$ \\
		\cmidrule{2-13}
		&\multirow{3}{*}{$48$} & $Ra_{1}$ & $96$ & $64.9_{2.8}$ & - & - & $96$ & $68.8_{11.0}$ & $96$ & $83.2_{1.2}$ & $96$ & $72.0_{6.5}$ \\
        && $Ra_{50}$ & $4,8k$ & $66.2_{1.5}$ & - & - & $4,8k$ & $74.4_{5.1}$ & $4,8k$ & $82.5_{4.3}$ & $4,8k$ & $77.5_{3.1}$\\
        && $IS$ & $1,2k$ & $66.9_{2.1}$ & - & - & $1,6k$ & \bftab{78.6$_{4.2}$} & $954$ & $87.0_{1.4}$ & $1.0k$ & $77.0_{4.6}$ \\
	    \cmidrule{2-13}
		&\multirow{3}{*}{$64$} & $Ra_{1}$ & $128$ & $65.5_{2.6}$ & - & - & $128$ & $72.4_{7.0}$ & $128$ & $84.0_{1.7}$ & $128$ & $74.4_{3.9}$\\
        && $Ra_{50}$ & $6,4k$ & $68.1_{1.5}$ & - & - & $6,4k$ & $76.1_{5.1}$ & $6,4k$ & $84.4_{2.5}$ & $6,4k$ & \bftab{78.6$_{3.1}$}\\
        && $IS$ & $1,6k$ & $67.7_{1.8}$ & - & - & $2,2k$ & $78.5_{5.2}$ & $1,3k$ & $87.7 _{1.6}$ & $1,3k$ & $77.7_{4.0}$\\
        \cmidrule{2-13}
		&\multirow{3}{*}{$128$} & $Ra_{1}$ & $256$ & $65.4_{2.9}$ & - & - & - & - & $256$ & $87.2_{2.2}$ & - & - \\
		&& $Ra_{50}$ & $12,8k$ & $68.7_{1.5}$ & - & - & - & - & $12,8k$ & $89.8_{1.2}$ & - & - \\
		&& $IS$ & $3,3k$ & $69.1_{1.8}$ & - & - & - & - & $2,6k$ & $90.5_{1.4}$ & - & -\\
		\cmidrule{2-13}
        &\multirow{3}{*}{$256$} & $Ra_{1}$ & $512$ & $66.6_{2.2}$ & - & - & - & - & $512$ & $89.3_{1.9}$ & - & - \\
		&& $Ra_{50}$ & $25,6k$ & $69.4_{1.2}$ & - & - & - & - & $25,6k$ & $95.4_{1.7}$ & - & - \\
		&& $IS$ & $5,2k$ & $69.6_{1.8}$ & - & - & - & - & $5,1k$ & $92.9_{1.4}$ & - & - \\
		\cmidrule{2-13}
	    &\multirow{3}{*}{$512$} & $Ra_{1}$ & $1,0k$ & $66.0_{2.6}$ & - & - & - & - & $1,0k$ & $91.7_{1.2}$ & - & - \\
		&& $Ra_{50}$ & $51,2k$ & \bftab{70.7$_{2.2}$} & - & - & - & - & $51,2k$ & \bftab{96.5$_{1.1}$} & - & - \\
		&& $IS$ & $13,0k$ & $70.0_{1.8}$ & - & - & - & - & $10,3k$ & $95.2_{0.8}$ & - & - \\
	     \bottomrule
	\end{tabular}}
	\caption{\label{tab:comparative_train_creation_instance_user_spanish} \textbf{Spanish Validation results} of the CA and SN models. \textit{n} shows the number of training users and \textit{s} the total training size. Top results are highlighted with \textbf{bold}.}
\end{table*}

\begin{table*} [!t]
	\centering
	\resizebox{1.0\textwidth}{!}{
	\begin{tabular}{p{1.9cm}p{0.6cm}p{0.85cm}p{0.85cm}p{0.85cm}p{0.85cm}p{0.85cm}p{0.85cm}p{0.85cm}p{0.85cm}p{0.85cm}p{0.85cm}p{0.85cm}p{0.85cm}}
	\toprule
		\multicolumn{14}{c}{\textbf{English dataset results}} \\
		\midrule
		&  & \multicolumn{2}{c}{Gender} & \multicolumn{2}{c}{Age} & 
		\multicolumn{2}{c}{Hate Speech} & \multicolumn{2}{c}{Bots} & \multicolumn{2}{c}{Fake News} & \multicolumn{2}{c}{Depression}\\
		\cmidrule(lr){3-4}    \cmidrule(lr){5-6}  \cmidrule(lr){7-8} \cmidrule(lr){9-10} \cmidrule(lr){11-12}    \cmidrule(lr){13-14} 
		System & \textit{n} & $Acc.$ & $F_{1}$ & $Acc.$ & $F_{1}$ & $Acc.$ & $F_{1}$ & $Acc.$ & $F_{1}$ & $Acc.$ & $F_{1}$ & $Acc.$ & $F_{1}$ \\
		\midrule
		winner        & all & \bftab{82.3} & - & $83.8$ & - & \bftab{74.0} & - & \bftab{96.0} & - & \bftab{75.0} & - & $41.3$ & - \\
        ST-lr         & all & $76.0$ & $76.0$  & $72.1$ & $47.4$ & $64.0$ & $63.8$ & $90.2$ & $90.2$ & $68.0$ & $67.8$ & $28.0$ & $20.2$ \\
		xlm-roberta   & all & $53.2$ & $39.5$  & $73.1$ & $41.9$ & $52.6$ & $39.0$ & $86.6$ & $86.3$ & $57.4$ & $47.4$ & $33.8$ & $21.3$  \\
		user-char-lr  & all & $79.2$ & $79.2$  & $73.9$ & $45.0$ & $62.4$ & $62.1$ & $91.2$ & $91.2$ & $70.5$ & $70.4$ & $35.2$ & $24.0$ \\
		LDSE          & all & $74.7$ & $74.7$  & \bftab{85.2} & \bftab{76.5} & $70.0$ & $70.0$ & $90.6$ & $90.5$ & $74.5$ & $74.5$ & \bftab{45.0} & \bftab{38.2}  \\
		\midrule
		CA      & $0$    & \bftab{75.3} & \bftab{75.2} & \bftab{63.4} & \bftab{35.1} & \bftab{70.0} & \bftab{69.9} & $52.3$ & $38.4$ & \bftab{64.0} & \bftab{63.9} & \bftab{36.2} & \bftab{25.5}\\
		SN & $0$    & $38.0$ & $37.6$ & $38.7$ & $29.7$ & $45.0$ & $44.1$ & \bftab{62.5} & \bftab{62.5} & $60.0$ & $56.3$ & $17.5$ & $ 15.9$\\
		\midrule
		CA      & all & \bftab{76.1} & \bftab{76.0} & $65.9$ & $36.6$ & \bftab{63.4} & \bftab{63.3} & \bftab{88.2} & \bftab{88.0} & $59.3$ & $56.2$ & \bftab{34.2} & $19.4$ \\
		SN & all  & $76.0$ & $76.0$ & \bftab{78.3} & \bftab{61.5} & $60.4$ & $60.2$ & $86.0$ & $85.8$ & \bftab{65.7} & \bftab{65.6} & $32.1$ & \bftab{22.9} \\
		\midrule
		CA   & best   & \bftab{77.3} & \bftab{77.3} & $72.3$ & $44.6$ & \bftab{70.0} & \bftab{69.9} & \bftab{87.8} & \bftab{87.7} & $62.6$ & $61.7$ & \bftab{33.5} & $24.7$\\
		SN & best & $76.3$ & $76.3$ & \bftab{72.4} & \bftab{61.2} & $62.2$ & $62.1$ & $87.4$ & $87.2$ & \bftab{65.5} & \bftab{65.1} & $29.8$ & \bftab{25.7}\\
		\toprule
		\multicolumn{14}{c}{\textbf{Spanish dataset results}} \\
		\midrule
		winner   &    all    & \bftab{83.2} & - & \bftab{79.6} & - & \bftab{85.0} & - & \bftab{93.3} & - & \bftab{82.0} & -  \\
        ST-lr    & all         & $70.3$ & $70.3$ & $62.7$ & $48.7$ & $80.8$ & $80.6$ & $87.3$ & $87.3$ & $76.5$ & $76.5$ \\
		xlm-roberta & all   & $53.8$ & $42.2$ & $50.0$ & $16.7$ & $75.6$ & $75.2$ & $86.2$ & $86.1$ & $71.4$ & $67.9$  \\
		user-char-lr   & all & $77.8$ & $77.8$ & $69.5$ & $56.5$ & $80.0$ & $79.9$ & $92.5$ & $92.5$ & $75.6$ & $75.4$  \\
		LDSE     &    all  & $71.9$ & $71.9$ & $78.4$ & $64.9$ & $82.0$ & $82.0$ & $83.7$ & $83.7$ & $79.0$ & $78.9$   \\
		\midrule
		CA    &    $0$   & \bftab{68.3} & \bftab{68.2} & $21.6$ & $16.4$ & \bftab{79.0} & \bftab{78.6} & \bftab{50.1} & $33.6$ & \bftab{51.0} & $36.3$\\
		SN & $0$ & $38.1$ & $33.9$ & \bftab{48.6} & \bftab{32.6} & $49.0$ & $34.5$ & $41.5$ & \bftab{38.4} &  $49.5$ & \bftab{44.3} \\
		\midrule
		CA      & all & \bftab{71.7} & \bftab{71.7} & $66.4$ & $51.1$ & $76.6$ & $76.4$ & \bftab{87.1} & \bftab{87.0} & \bftab{78.3} & \bftab{78.0}\\
		SN & all  & $71.0$ & $71.0$ & \bftab{66.8} & \bftab{51.6} & \bftab{77.4} & \bftab{77.1} & $83.6$ & $83.2$ & $77.8$ & $77.6$ \\
		\midrule
		CA   & best      & \bftab{73.7} & \bftab{73.4} & $40.9$ & $27.8$ & $80.0$ & $79.8$ & \bftab{88.3} & \bftab{88.2} & $74.7$ & $74.4$\\
		SN & best & $72.7$ & $72.5$ & \bftab{66.8} & \bftab{56.0}& \bftab{80.8} & \bftab{80.7}& $86.5$ & $86.4$& \bftab{76.3} & \bftab{76.2} \\
		\bottomrule
	\end{tabular}}
		\caption{\label{tab:test_set_results} \textbf{Test set results} of CA and SN compared to several baseline and reference approaches. \textit{n} shows the number of training users. Per-block top results without confidence interval overlaps are highlighted with \textbf{bold}.}
\end{table*}


In this section, we analyze the results of the Siamese networks (SN) and the Cross-Attention (CA) models among different author profiling tasks in English and Spanish. 
We compare the performance in function of the number of training users ($n$), author instance selection method (Inst. sel.), and total training size ($s$). 
The few-shot results of user-char-lr can be found in the Appendix (S. \ref{sec:appendix}).

Table \ref{tab:comparative_train_creation_instance_user_english} shows the English validation results.
Regarding the zero-shot setting ($n=0$), SN outperforms CA in Age and Bots.
However, it obtains lower numbers in tasks such as Gender and Hate Speech.
Regarding the few-shot setting ($n>0)$, the $Ra_{50}$ author instance selection method outperforms $Ra_{1}$ in combination with SN.
Interestingly, the contrary happens for CA, where $Ra_{50}$ is only superior in Bots.
Comparing $Ra_x$ and $IS$, with SN both obtain similar results in four tasks and the later exceeds in two. 
For the CA model, $IS$ is superior in three tasks and similar in the rest.
$IS$ uses much less training data ($s$) than $Ra_{50}$ in all cases and thus reduces training time and cost.
Looking at the overall English few-shot results, SN outperforms CA in four tasks (Age, Hate Speech, Fake News and Depression) and is out-performed in two (Gender and Bots).
However, there exist overlaps between some confidence intervals, suggesting a low significance.
SN gives more stable results across tasks.
Finally, there is a general trend of more training users giving better results.
Nevertheless, as proved by our $IS$ method the information used from those users matters.


Table \ref{tab:comparative_train_creation_instance_user_spanish} shows the Spanish validation results. 
Regarding the zero-shot setting, similarly to English, SN outperforms CA in Age and Bots.
In Addition, it also improves in Fake News.
Looking at the few-shot setting, the results also show a clear improvement trend while increasing the number of training users.
This is more clear for SN, improving from the beginning ($n=8$), than for CA, which sometimes requires additional training users and failed at improving the Age zero-shot results.
Regarding the author instance selection method, in general, \textit{IS} in combination with CA outperforms random selection on most datasets.
For SN, we find similar results for $Ra_{50}$ and $IS$, while $IS$ uses  ~75\% less training instances. 
This highlights again its capability to select the most relevant training instances.
Looking at the overall Spanish few-shot results, SN outperforms in three tasks (Age, Hate Speech and Fake News) and CA in two (Gender and Bots).
However, both models offer a similar performance;  with the exception of Age, where CA failed to converge.
This, together with the English results, leave SN as the most stable across tasks and languages.



The few-shot experiments with the test sets use the best configurations obtained at our validation phase, i.e., the number of training users ($n=\text{best}$) and author instance selection method for each language and task. 
Table \ref{tab:test_set_results} compares the test set results of CA and SN with several baselines and reference approaches (see Section \ref{sec:baselines}). 
Note that the baselines use all the available training data ($n=\text{all}$) and that we show both the SN and CA zero and few-shot results.\footnote{The standard deviation results of Table \ref{tab:test_set_results} are omitted due to space. However, this does not affect our analysis. Omitted values can be found in Appendix (S. \ref{sec:appendix}).}
As you can see, the zero-shot models outperform xlm-roberta in English.
CA also does it for Spanish.
Comparing them against other approaches, including the shared task top systems ($winner$), we find encouraging results: the zero-shot models ranked third at some tasks, e.g. EN CA Hate Speech, EN CA Gender, and EN CA Depression.
This is remarkable considering that those models did not use any training data, and shows their potential for low-resource author profiling.
Note that few shot (\textit{best}) worked similarly or better than the CA and SN models trained with all the train set (\textit{all}).
Interestingly, LDSE, a popular profiling baseline, ranks first for Depression.\footnote{The accuracy of the eRisk SOTA system corresponds to the DCHR metric used in the shared task.}

As expected, few-shot out-performs zero-shot in most tasks with few exceptions for CA: EN Hate Speech, EN Fake News, and EN Depression.
Similar to the validation experiments, as CA failed to converge for some tasks, SN offers higher stability.
Finally, the comparison of the few-shot models with the rest of approaches shows it is ranking third for some tasks and languages: EN CA Gender, EN CA Hate Speech, ES CA Bots, ES SN Fake News.
In comparison with \emph{winner} the FSL models reach 80\% of the accuracy of the winner system in 90\% of the cases and 90\% in 50\% of the cases. They do so using less than 50\% of the training data. This shows that FSL often yields competitive systems with less individual tuning and less annotation effort. 
\section{Conclusions}
\label{sec:conclusions}

In this work, we studied author profiling from a low-resource perspective: with little or no training data. 
We addressed the task using zero and few-shot text classification.
We studied the identification of age, gender, hate speech spreaders, fake news spreaders, bots, and depression.
In addition, we analyzed the effect of the entailment hypothesis and the size of the few-shot training sample on the system's performance.
We evaluated corpora both in Spanish and English.

On the comparison of Cross Attention and Siamese networks, we find that the former performs better in the zero-shot scenario while the latter 
gives more stable few-shot results across the evaluated tasks and languages.
We find that entailment-based models out-perform supervised text classifiers based on roberta-XLM
and that we can reach 80\% of the state-of-the-art accuracy using less than 50\% of the training data on average.
This highlights their potential for low-resource author profiling scenarios.

\section{Appendix}
\label{sec:appendix}

The repository at \url{https://tinyurl.com/ZSandFS-author-profiling} contains experimental details and results.

\section*{Acknowledgements}
We gracefully thank the support of the Pro$^2$Haters - Proactive Profiling of Hate Speech Spreaders (CDTi IDI-20210776), DETEMP - Early Detection of Depression Detection in Social Media (IVACE IMINOD/2021/72) and DeepPattern (PROMETEO/2019/121) R\&D grants. Grant PLEC2021-007681 funded by MCIN/AEI/ 10.13039/501100011033 and by European Union NextGenerationEU/PRTR.



\bibliography{references_bak}
\end{document}